\documentclass[letterpaper, 10 pt, conference]{ieeeconf}

\IEEEoverridecommandlockouts
\usepackage[hyphens]{url}  
\usepackage{graphicx} 
\usepackage{caption} 
\usepackage{algorithm}
\usepackage{algorithmic}
\usepackage{multirow}
\usepackage{amsmath,amssymb,amsfonts}

\usepackage{newfloat}
\usepackage{listings}
\DeclareCaptionStyle{ruled}{labelfont=normalfont,labelsep=colon,strut=off} 
\floatstyle{ruled}
\newfloat{listing}{tb}{lst}{}
\floatname{listing}{Listing}

\usepackage{booktabs}

\title{Hierarchical Prototype-Memory Adaptation of SAM for Surgical Instrument Segmentation}
\author{
    Xinning Yao$^{1}$, 
    Jingjing Wang$^{1}$, 
    Jinghua Yue$^{1}$,
    Xiaoyan Luo$^{1}$,
    Fugen Zhou$^{1}$,
    Bo Liu$^{1}$
\thanks{$^{1}$Image Processing Center, Beihang University, Beijing, 100191, China.  Corresponding Author: {\tt\small bo.liu@buaa.edu.cn}}
}

\begin{document}

\maketitle
\begin{abstract}
Surgical instrument segmentation (SIS) is fundamental for computer-assisted surgery, where reliable instrument masks enable precise scene understanding and clinical assistance. Recently, adapting foundation models like the Segment Anything Model (SAM) to the surgical domain via prompt-learning has shown encouraging results. However, the performance of these adapted models under challenging surgical conditions is constrained by suboptimal adaptation mechanisms. Specifically, optimizing prompts or prototypes purely via downstream segmentation loss tends to cause them to degenerate into task-specific parameters rather than serving as persistent, stable category memory, thereby degrading their robustness against complex intraoperative variations. Moreover, routing multi-scale visual cues through a single prompt pathway creates a bottleneck that hinders effective scale-matched coupling. To address these limitations, we propose HPMA, a Hierarchical Prototype-Memory Adaptation framework for SAM. Specifically, HPMA constructs a frozen, multi-scale visual prototype memory bank from annotated surgical scenes and integrates it into SAM's feature space using lightweight adapters to preserve stable category evidence. To maximize the utility of multi-scale cues, we introduce a scale-matched coupling mechanism where global prototypes calibrate class-level prompt features, structural prototypes guide decoder object queries, and local prototypes align high-resolution feature maps through a local alignment objective. Extensive experiments on the public EndoVis2017 and EndoVis2018 datasets demonstrate that our approach achieves state-of-the-art performance, outperforming existing foundation model adaptation methods.
\end{abstract}


\section{Introduction}

Accurate surgical instrument segmentation (SIS) plays a fundamental role in computer-assisted surgery, where reliable instrument masks are essential for precise scene understanding and clinical assistance. In recent years, vision foundation models, represented by the Segment Anything Model (SAM) and its multimodal evolution SAM3 \cite{kirillov2023segment, ravi2025sam, carion2026sam}, have demonstrated promising potential in general-purpose segmentation tasks. However, directly applying these models to the surgical domain is hindered by a substantial domain gap. Unlike objects in natural images, surgical instrument categories must be recognized from highly domain-specific visual evidence. Instruments often exhibit delicate structures, while surgical scenes commonly involve strong specular reflections, mutual occlusions, and ambiguous boundaries \cite{ma2025medsam2, zhao2025rethinking}. Consequently, conventional parameter fine-tuning methods are often insufficient to bridge this gap effectively.

\begin{figure}
\centerline{\includegraphics[width=0.95\linewidth]{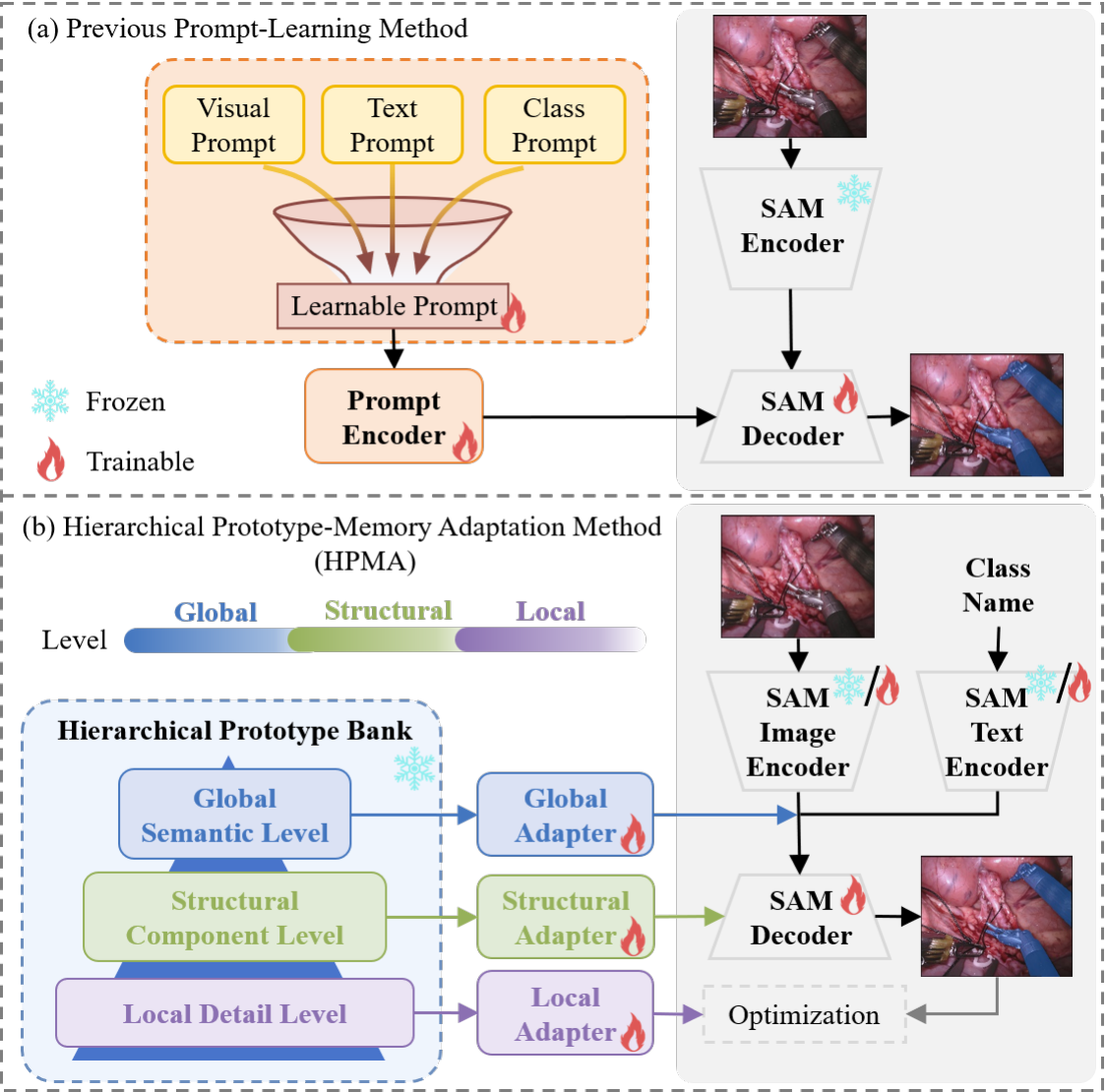}}
\caption{The conceptual comparison of (a) previous prompt-learning method and our (b) hierarchical prototype-memory adaptation method.}
\label{fig:compare}
\end{figure}

To address this challenge, existing prompt-learning methods \cite{liu2025resurgsam2, tang2026distillation}, such as SurgicalSAM \cite{yue2024surgicalsam}, have attempted to incorporate domain-specific knowledge into foundation models, demonstrating their value in mitigating the discrepancy between generic representations and surgical semantics. Nevertheless, their performance under complex and highly variable surgical conditions remains constrained by suboptimal adaptation mechanisms. Specifically, existing methods suffer from two major limitations, as shown in Fig. \ref{fig:compare}(a). First, optimizing prompts or prototypes exclusively via downstream segmentation objectives frequently causes them to overfit as task-specific parameters, eroding their capacity to act as persistent, stable category memories and diminishing model robustness against intraoperative variations \cite{tang2026distillation, Lu_2026_CVPR}. Second, existing approaches typically route multi-scale visual cues through a single prompting pathway. This design creates a representational bottleneck that prevents effective scale-matched coupling between visual features and the corresponding model components \cite{yue2023surgicalpart, Zhu_2026_CVPR}.

To overcome these limitations, we propose Hierarchical Prototype-Memory Adaptation (HPMA), a novel adaptation framework for SAM. To address the first limitation of prompt degeneration, HPMA explicitly decouples domain knowledge storage from downstream feature adaptation. As shown in Fig. \ref{fig:compare}, instead of optimizing prompts blindly, HPMA extracts visual evidence from annotated surgical scenes to construct a frozen multi-scale prototype memory bank. By clustering and preserving diverse visual evidence of surgical instruments, the memory bank serves as a persistent statistical memory insulated from downstream optimization objectives. We then introduce lightweight adapters that learn to transform this stable, class-specific visual evidence into effective feature enhancement, seamlessly injecting them into the feature space of SAM.

Furthermore, we argue that surgical instrument segmentation is not merely a single-scale adaptation problem. A model should capture not only the overall appearance of an instrument but also its complex internal structures and fine-grained details. Accordingly, visual evidence at different scales should be delivered to the model components best suited to exploit it. To this end, we design a hierarchical representation level-to-module coupling mechanism. At the global semantic level, global prototypes calibrate class-level text-prompt features before feature fusion. At the structural component level, structural prototypes are injected into the Transformer decoder to guide and refine object queries, enabling the model to better capture complex instrument structures. At the local detail level, local prototypes serve as local alignment targets for high-resolution feature maps during training, effectively sharpening fine-grained instrument details without introducing additional computational overhead at inference time.

In summary, our main contributions are as follows:
\begin{itemize}
\item We propose HPMA, a framework that explicitly models visual evidence from the surgical domain as a prototype memory. Through lightweight adapters, HPMA transforms this frozen visual memory into model-specific feature enhancement, improving segmentation robustness in complex surgical scenes.

\item We develop a hierarchical representation level-to-module coupling mechanism that reduces the representational bottleneck imposed by a single adaptation pathway. By adapting prototype memory at different scales to the appropriate stages of the model, the proposed mechanism more effectively alleviates class-level semantic mismatch, insufficient structural perception, and fine-grained detail deficiencies in surgical instrument segmentation.

\item We conduct extensive experiments on two public benchmarks, EndoVis2017 and EndoVis2018. The results demonstrate that HPMA substantially outperforms existing state-of-the-art methods and establishes new benchmark performance in terms of Challenge IoU and mean class IoU, with minimal inference overhead compared to existing adaptation methods.
\end{itemize}

\section{Related Work}
\subsection{Surgical Instrument Segmentation}
Surgical instrument segmentation is challenging due to specular reflections, occlusions, ambiguous boundaries, and high intra-class variations. Specialized methods have tackled these issues via mask-attended classification \cite{baby2023forks}, masked-attention transformers with temporal modeling \cite{ayobi2023matis}, vision-language representation \cite{rao2022denseclip, zhou2023text}, or instance-level representation learning \cite{sestini2025saf}. Despite their effectiveness, these specialist models encode surgical knowledge implicitly in task-specific parameters, limiting their robustness against complex and changing intraoperative conditions.

\subsection{SAM Adaptation for Medical and Surgical Segmentation}

The domain gap between natural and medical images has driven extensive adaptation of the Segment Anything Model (SAM) \cite{kirillov2023segment,ravi2025sam,carion2026sam}. Methods like MedSAM \cite{ma2024segment} perform full fine-tuning, whereas SAMed \cite{zhang2023samed} and Medical SAM Adapter \cite{wu2025medical} introduce LoRA or lightweight adapters for parameter-efficient adaptation. Other variants tune specialized components via text guidance \cite{paranjape2024adaptivesam}, hierarchical decoding \cite{cheng2024unleashing}, or decoder adaptation prompts \cite{tejero2025sam,yang2025rsg}. However, domain knowledge in these approaches is predominantly encoded in optimized model parameters, causing the degeneration of foundation model priors and leaving models susceptible to semantic drift under challenging surgical artifacts.

\begin{figure*}
\centerline{\includegraphics[width=0.98\textwidth]{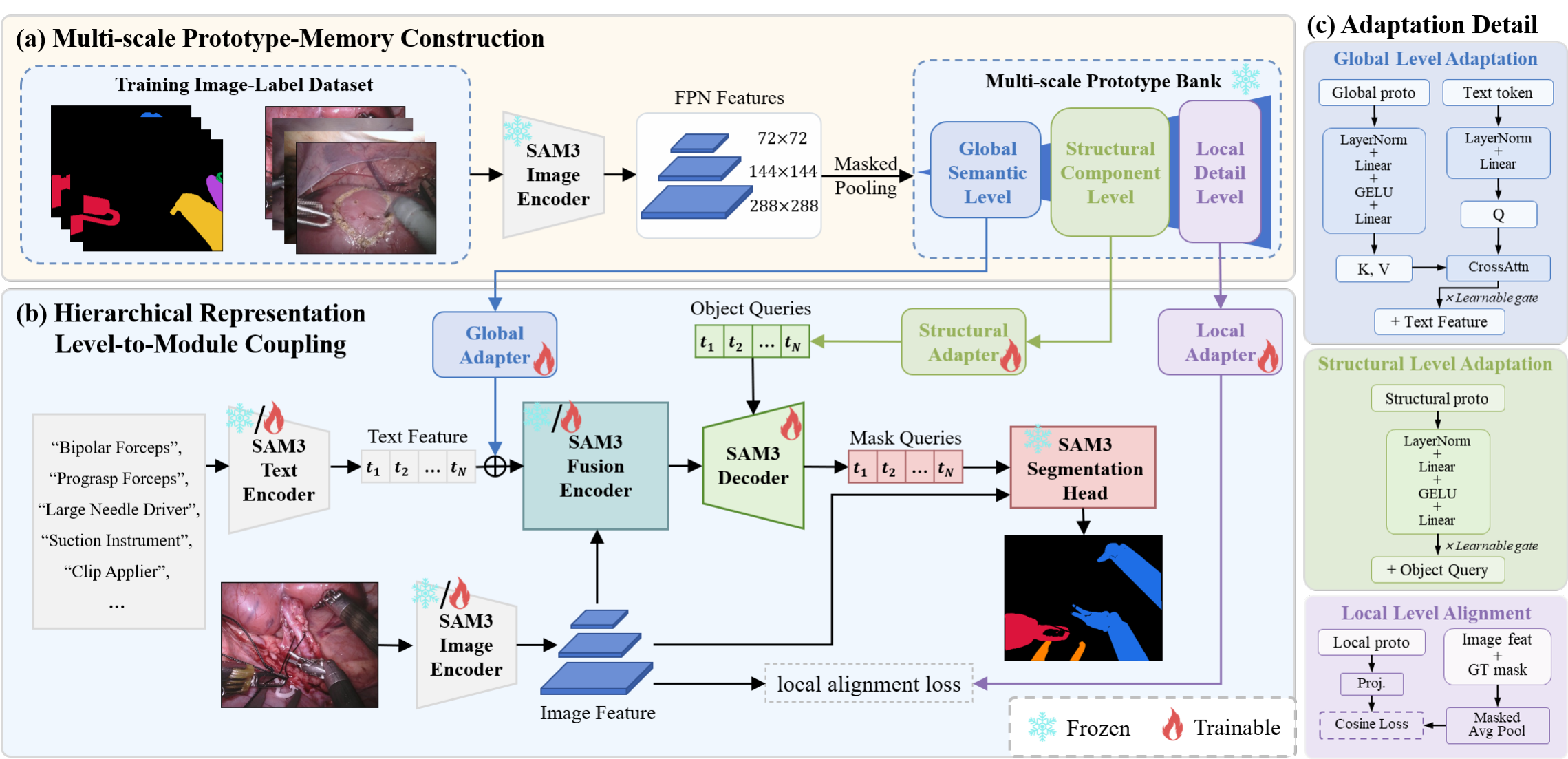}}
\caption{Overview of the proposed HPMA framework. (a) The Multi-scale Prototype-Memory Construction extracts multi-scale FPN features from labeled surgical images using a frozen SAM3 image encoder and constructs global-, structural-, and local-level prototype banks. (b) The Hierarchical Representation Level-to-Module Coupling mechanism injects global prototypes into text features before multimodal fusion, uses structural prototypes to refine decoder object queries, and aligns high-resolution image features with local prototypes during training. (c) The Adaptation Detail of the modules at each level.}
\label{fig:overview}
\end{figure*}

\subsection{Prompt Learning for Surgical Segmentation}

Prompt learning reduces manual interaction in SAM adaptation. AutoSAM \cite{shaharabany2023autosam} and Surgical-DeSAM \cite{sheng2024surgical} automatically generate prompts via auxiliary encoders or detectors. SurgicalSAM \cite{yue2024surgicalsam} and SurgicalPart-SAM \cite{yue2023surgicalpart} introduce prototype-based or part-level prompts, while other works leverage hierarchical prompt trees \cite{Zhu_2026_CVPR} or prompt distillation \cite{tang2026distillation}. Nevertheless, existing prompt-learning approaches route multi-scale visual cues through a single, congested prompting pathway, creating a representational bottleneck. In contrast, HPMA explicitly preserves multi-scale visual evidence in a frozen prototype memory and couples each representation level with its optimal model component.

\section{Methodology}

\subsection{Problem Formulation and Overall Architecture}

We formulate surgical instrument segmentation as a class-prompted binary mask prediction task. Given a surgical image $I \in \mathbb{R}^{H \times W \times 3}$ and a text prompt $t_c$ for category $c \in \mathcal{C}$, the network $f_\Theta$ predicts a binary mask $\hat{Y}_c = f_\Theta(I, t_c) \in \{0, 1\}^{H \times W}$, where $\Theta = \{\Theta_{\text{frozen}}, \Theta_{\text{adapter}}\}$ separates frozen foundation parameters from trainable adapters. Multi-class masks are obtained by querying each category independently and taking the element-wise maximum.

We adopt SAM3 \cite{carion2026sam} as our baseline, which comprises an image encoder $E_{\text{img}}$, text encoder $E_{\text{txt}}$, fusion encoder $E_{\text{fus}}$, transformer decoder $D_{\text{trans}}$, and segmentation head $H_{\text{seg}}$. Native SAM3 extracts multi-scale visual features $F = \{F_l\}_{l=1}^3$ and text embeddings $T_c$, fuses them via $E_{\text{fus}}$, and predicts masks using decoder object queries $Q$.

As shown in Fig. 2, HPMA introduces a frozen prototype memory bank $\mathcal{P}$ paired with lightweight adapters across three hierarchical levels: (1) Global level: Calibrating raw text features $T_c$ with global prototypes $\mathcal{P}_c^g$ into $\tilde{T}_c$ before fusion; (2) Structural level: Modifying object queries $Q$ with structural prototypes $\mathcal{P}_c^p$ into $\tilde{Q}_c$; and (3) Local level: Aligning high-resolution features $F_0$ with local prototypes $\mathcal{P}_c^l$ via a training-only objective. Formally, given intermediate features $(F, T_c) = (E_{\text{img}}(I), E_{\text{txt}}(t_c))$, the refined execution flow is:
\begin{equation}
\hat{Y}_c = H_{\text{seg}}\Big(D_{\text{trans}}\big(E_{\text{fus}}(F, \tilde{T}_c), \tilde{Q}_c\big)\Big).
\label{eq:1}
\end{equation}

\subsection{Construction and Adaption of Multi-scale Prototype-Memory}
The core of our framework is the construction of a frozen visual prototype memory bank and the design of lightweight adapters that inject this evidence into the model's backbone to enhance feature representations.

To construct a stable repository of category-specific evidence, we extract multi-scale feature maps from the Feature Pyramid Network (FPN) of the frozen SAM3 image encoder. For each representation scale $s \in \{\text{global, structural, local}\}$ and instrument category $c \in \mathcal{C}$, we aggregate image-level features via masked average pooling governed by the ground-truth binary masks $M_c$. To encapsulate the severe intra-class visual variations inherent to surgical scenes, these aggregated representations are subsequently partitioned offline into $K$ distinct prototype vectors via K-Means clustering prior to training, formulating the comprehensive memory bank $\mathcal{P} \in \mathbb{R}^{3 \times \vert{}\mathcal{C}\vert{} \times K \times d}$.

To facilitate rapid and parameter-efficient adaptation, we design lightweight adapters that read from the frozen memory bank and inject visual evidence into the active optimization pathway. The bank $\mathcal{P}$ remains strictly frozen to prevent the prototypes from degenerating into task-specific parameters during downstream optimization, avoiding catastrophic forgetting of SAM3's foundational priors. For a given scale $s$, the adapter takes a target SAM3 representation $Z_c^s$ and computes a residual update using a lightweight mapping function $\Psi_s$:
\begin{equation}
\tilde{Z}_c^s = Z_c^s + \alpha_s \Psi_s(Z_c^s, \mathcal{P}_c^s),
\label{eq:2}
\end{equation}
where $\alpha_s$ is a learnable scalar initialized to a small value. This formulation explicitly isolates evidence storage from feature adaptation: the memory dictates what an instrument looks like under varying conditions, while the adapter learns how to optimally integrate this evidence into the backbone.

\subsection{Hierarchical Representation Level-to-Module Coupling}

Unlike methods constrained by a singular projection bottleneck, our approach capitalizes on the rich, three-tiered hierarchical FPN features. Inspired by \cite{Zhu_2026_CVPR, yue2023surgicalpart}, we introduce scale-matched adapters to explicitly route these visual cues into the most receptive SAM3 modules, maximizing representational efficiency. Specifically, the global level addresses category appearance, the structural level guides object-query formation, and the local level refines high-resolution features.

At the global semantic level, feature maps capture the overall context of the instrument. We design a global adapter to inject this evidence into class prompts prior to image-text fusion. Acting as a cross-attention mechanism with text tokens $T_c$ as queries, we define $Q_c = T_c W_Q$, $K_c = \Phi_g(\mathcal{P}_c^g) W_K$, and $V_c = \Phi_g(\mathcal{P}_c^g) W_V$, and the update is formulated as:
\begin{equation}
A_c^g = \mathrm{softmax} \left( \frac{Q_c K_c^\top}{\sqrt{d_k}} \right), \quad \widetilde{T}_c = T_c + \alpha_g (A_c^g V_c) W_O,
\label{eq:text_enhanced}
\end{equation}
where $W_{\{Q,K,V,O\}}$ are learnable projection matrices, $d_k$ is the scaling dimension, and $\alpha_g$ is a learnable scalar. The projection $\Phi_g$ comprises a LayerNorm and a two-layer MLP with GELU. This formulation explicitly anchors abstract text embeddings into the domain-specific surgical visual manifold ($\mathcal{P}_c^g$) to overcome modality gaps from intraoperative variations, ensuring strictly domain-aligned representations for the fusion encoder.

At the intermediate structural level, geometric sub-components of the instruments are represented. We inject this information directly into the SAM3 transformer decoder by modifying the learnable object queries $Q$. Since these queries function as spatial anchors, adding the structural component prototype endows them with the geometric bias of specific instrument parts. A projection $\Phi_p$, structurally identical to $\Phi_g$, maps the structural prototype $\mathcal{P}_c^p$ into a query residual:
\begin{equation}
\widetilde{Q}_c = Q + \alpha_p \Phi_p(\mathcal{P}_c^p).
\label{eq:query_enhanced}
\end{equation}
Rather than relying on generic initialization, this explicitly biases queries toward inherent instrument structures, which is crucial for highly articulated tools defined by distinct components like shafts, jaws, and tips.

At the finest granularity, local-level features encode critical boundaries and textures. To exploit this without inference latency, we formulate a local alignment objective during training. The high-resolution feature map $F_0$ is spatially pooled over the target object region to yield a compact representation:
\begin{equation}
\phi_i = \frac{\sum_{x} w_i(x) F_0(x)}{\sum_{x} w_i(x) + \epsilon},
\label{eq:local_pooling}
\end{equation}
where $w_i(x) \in \{0, 1\}$ denotes the resized ground-truth mask value at spatial coordinate $x$. This is then explicitly regularized by aligning it toward the corresponding local prototype $\mathcal{P}_c^l$.

By systematically aggregating information across these dimensions, HPMA circumvents the representational bottleneck of single-pathway prompting. This hierarchical coupling seamlessly bridges the perceptual gap---from semantic calibration to structural awareness and local detail refinement---comprehensively improving multi-class segmentation capabilities.

\begin{table*}[t]
\centering
\caption{Comparative Results on the EndoVis2017 Dataset. The best results are presented in bold.}
\resizebox{\textwidth}{!}{
\begin{tabular}{lcccccccccccc}
\toprule
\multirow{2}{*}{Method} & \multirow{2}{*}{Ref.} & \multirow{2}{*}{Challenge IoU} & \multirow{2}{*}{IoU} & \multirow{2}{*}{mc IoU} & \multirow{2}{*}{mDice} & \multicolumn{7}{c}{Instrument Categories} \\
\cmidrule(lr){7-13}
& & & & & & BF & PF & LND & VS & GR & MCS & UP \\
\midrule
ISINet & MICCAI’ 20 & 55.62 & 52.20 & 28.96 & 52.00 & 38.70 & 38.50 & 50.09 & 27.43 & 2.01 & 28.72 & 12.56 \\
MATIS Frame & ISBI’ 23 & 68.79 & 62.74 & 37.30 & 53.06 & 66.18 & 50.99 & 52.23 & 32.84 & 15.71 & 19.27 & 23.90 \\
S3Net & WACV’ 23 & 72.54 & 71.99 & 46.55 & 66.44 & 75.08 & 54.32 & 61.84 & 35.50 & 27.47 & 43.23 & 28.38 \\
SCI-Net & JBHI’ 26 & 72.26 & 69.08 & - & - & - & - & - & - & - & - & - \\
\hline
TrackAnything & - & 67.41 & 64.50 & 62.97 & - & 55.42 & 44.46 & 62.43 & 67.03 & 65.17 & 83.68 & 62.59 \\
PerSAM & - & 42.47 & 42.47 & 41.80 & - & 53.99 & 25.89 & 50.17 & 52.87 & 24.24 & 47.33 & 38.16 \\
SurgicalSAM & AAAI’ 24 & 69.94 & 69.94 & 67.03 & 80.34 & 68.30 & 51.77 & 75.52 & 68.24 & 57.63 & \textbf{86.95} & 60.80 \\
MA-SAM2 & MICCAI’ 25 & 62.49 & 62.49 & 59.89 & - & 54.41 & 50.41 & 64.73 & 73.72 & 32.66 & 72.64 & 70.85 \\
Distillation-SAM & TMI' 26 & - & - & 75 & 80 & - & - & - & - & - & - & - \\
Ours & - & \textbf{79.63} & \textbf{79.63} & \textbf{76.34} & \textbf{83.87} & \textbf{83.71} & \textbf{69.26} & \textbf{87.25} & \textbf{83.13} & \textbf{48.52} & 80.32 & \textbf{75.02} \\
\hline
GT Centroid + SAM3 & ICLR’ 26 & 50.52 & 50.52 & 50.60 & 53.56 & 51.96 & 43.64 & 39.18 & 46.80 & 42.66 & 75.18 & 66.87 \\
\bottomrule
\end{tabular}
}

\label{tab:endovis2017}
\end{table*}

\begin{table*}[h]
\centering
\caption{Comparative Results on the EndoVis2018 Dataset. The best results are presented in bold.}
\resizebox{\textwidth}{!}{
\begin{tabular}{lcccccccccccc}
\toprule
\multirow{2}{*}{Method} & \multirow{2}{*}{Ref.} & \multirow{2}{*}{Challenge IoU} & \multirow{2}{*}{IoU} & \multirow{2}{*}{mc IoU} & \multirow{2}{*}{mDice} & \multicolumn{7}{c}{Instrument Categories} \\
\cmidrule(lr){7-13}
& & & & & & BF & PF & LND & MCS & UP & SI & CA \\
\midrule
ISINet & MICCAI’ 20 & 73.03 & 70.97 & 40.21 & 71.10 & 73.83 & 48.61 & 30.98 & 88.16 & 2.16 & 37.68 & 0.00 \\
MATIS Frame & ISBI’ 23 & 82.37 & 77.01 & 48.65 & 63.66 & 83.35 & 38.82 & 40.19 & \textbf{93.18} & 16.17 & 64.49 & 4.32 \\
S3Net & WACV’ 23 & 75.81 & 74.02 & 42.58 & 68.02 & 77.22 & 50.87 & 19.83 & 92.12 & 7.44 & 50.59 & 0.00 \\
SCI-Net & JBHI’ 26 & 83.59 & 81.10 & - & - & - & - & - & - & - & - & - \\
\hline
TrackAnything & - & 65.72 & 60.88 & 38.60 & - & 72.90 & 31.07 & 64.73 & 61.05 & 17.93 & 10.24 & 12.28 \\
PerSAM & - & 49.21 & 49.21 & 34.55 & - & 51.26 & 34.40 & 46.75 & 52.28 & 25.62 & 16.45 & 15.07 \\
SurgicalSAM & AAAI’ 24 & 80.33 & 80.33 & 58.87 & 69.61 & \textbf{83.66} & 65.63 & 58.75 & 88.56 & 21.23 & 54.48 & 39.78 \\
MA-SAM2 & MICCAI’ 25 & 64.40 & 64.40 & 62.13 & - & 56.93 & 44.91 & 66.73 & 75.35 & 74.51 & 36.82 & 37.65 \\
Ours & - & \textbf{84.12} & \textbf{84.12} & \textbf{77.20} & \textbf{87.43} & 80.69 & \textbf{81.29} & \textbf{78.23} & 92.12 & \textbf{67.73} & \textbf{80.34} & \textbf{60.04} \\
\hline
GT Centroid + SAM3 & ICLR’ 26 & 60.45 & 60.45 & 58.39 & 51.32 & 42.56 & 51.42 & 47.60 & 82.93 & 62.97 & 51.94 & 69.29 \\
\bottomrule
\end{tabular}
}

\label{tab:endovis2018}
\end{table*}

\subsection{Training Objective}

The overall training objective integrates the standard SAM3 loss formulation with our proposed local prototype alignment term. The base loss, denoted as $\mathcal{L}_{\mathrm{SAM3}}$, comprises focal (for mask classification), dice, bounding box, generalized IoU, category classification, and presence supervision losses:
\begin{equation}
\begin{split}
\mathcal{L}_{\mathrm{SAM3}} &= \lambda_{\mathrm{mask}}\mathcal{L}_{\mathrm{mask}} + \lambda_{\mathrm{dice}}\mathcal{L}_{\mathrm{dice}} + \lambda_{\mathrm{box}}\mathcal{L}_{\mathrm{box}} \\
&\quad + \lambda_{\mathrm{giou}}\mathcal{L}_{\mathrm{giou}} + \lambda_{\mathrm{cls}}\mathcal{L}_{\mathrm{cls}} + \lambda_{\mathrm{pres}}\mathcal{L}_{\mathrm{pres}},
\label{eq:base_loss}
\end{split}
\end{equation}
where the $\lambda$ coefficients balance the respective loss components.

To supervise the high-resolution features, we formulate a local alignment loss $\mathcal{L}_{\mathrm{align}}$ based on a cosine distance metric. We explicitly utilize cosine similarity to optimize the directional alignment of features rather than their $L_2$ magnitude. This design choice renders the high-resolution representations invariant to absolute feature scaling and highly resilient to illumination variations, significantly improving the model's robustness when delineating fine-grained edges amidst strong surgical specular reflections. 

For a mini-batch containing $N$ target object regions, the loss is computed between the spatially pooled high-resolution feature $\phi_i$ and its most semantically relevant projected local prototype $p_{c_i, k}^{l}$ (where $k \in \{1, \dots, K\}$ indexes the prototypes for the ground-truth category $c_i$):
\begin{equation}
\mathcal{L}_{\mathrm{align}} = \frac{1}{N} \sum_{i=1}^N \min_{k \in \{1, \dots, K\}} \left(1 - \cos(\phi_i, p_{c_i, k}^{l})\right). 
\label{eq:align_loss}
\end{equation}

The complete optimization objective is formulated as:
\begin{equation}
\mathcal{L} = \mathcal{L}_{\mathrm{SAM3}} + \lambda_{\mathrm{align}}\mathcal{L}_{\mathrm{align}},
\label{eq:total_loss}
\end{equation}
where $\lambda_{\mathrm{align}}$ is the weighting hyperparameter for the local alignment regularization.

\section{Experiments}

\subsection{Datasets and Evaluation Metrics}

We evaluate our proposed HPMA on two widely adopted public benchmarks for surgical instrument segmentation: EndoVis2017 and EndoVis2018.

\textbf{EndoVis2017} \cite{allan20192017} contains 8 videos for cross-validation, featuring 7 distinct instrument categories: Bipolar Forceps (BF), Prograsp Forceps (PF), Large Needle Driver (LND), Vessel Sealer (VS), Grasping Retractor (GR), Monopolar Curved Scissors (MCS), and Ultrasound Probe (UP). We use the 4-fold cross-validation provided by Shvets et al. \cite{shvets2018automatic}.

\textbf{EndoVis2018} \cite{allan20202018} comprises 11 training videos and 4 validation videos. It also evaluates 7 categories, substituting Vessel Sealer and Grasping Retractor with Suction Instrument (SI) and Clip Applier (CA). We use the instrument-type segmentation annotation given by González et al. \cite{gonzalez2020isinet}. 

To comprehensively evaluate the performance, we utilize four standard evaluation metrics \cite{gonzalez2020isinet, tang2026distillation}: Challenge IoU, IoU, mean class IoU (mc IoU) and mean Dice (mDice). Challenge IoU is a specialized metric that evaluates intersection over union exclusively for the instrument categories actively visible in a given frame.

\begin{figure*}[!t]
\centerline{\includegraphics[width=0.8\textwidth]{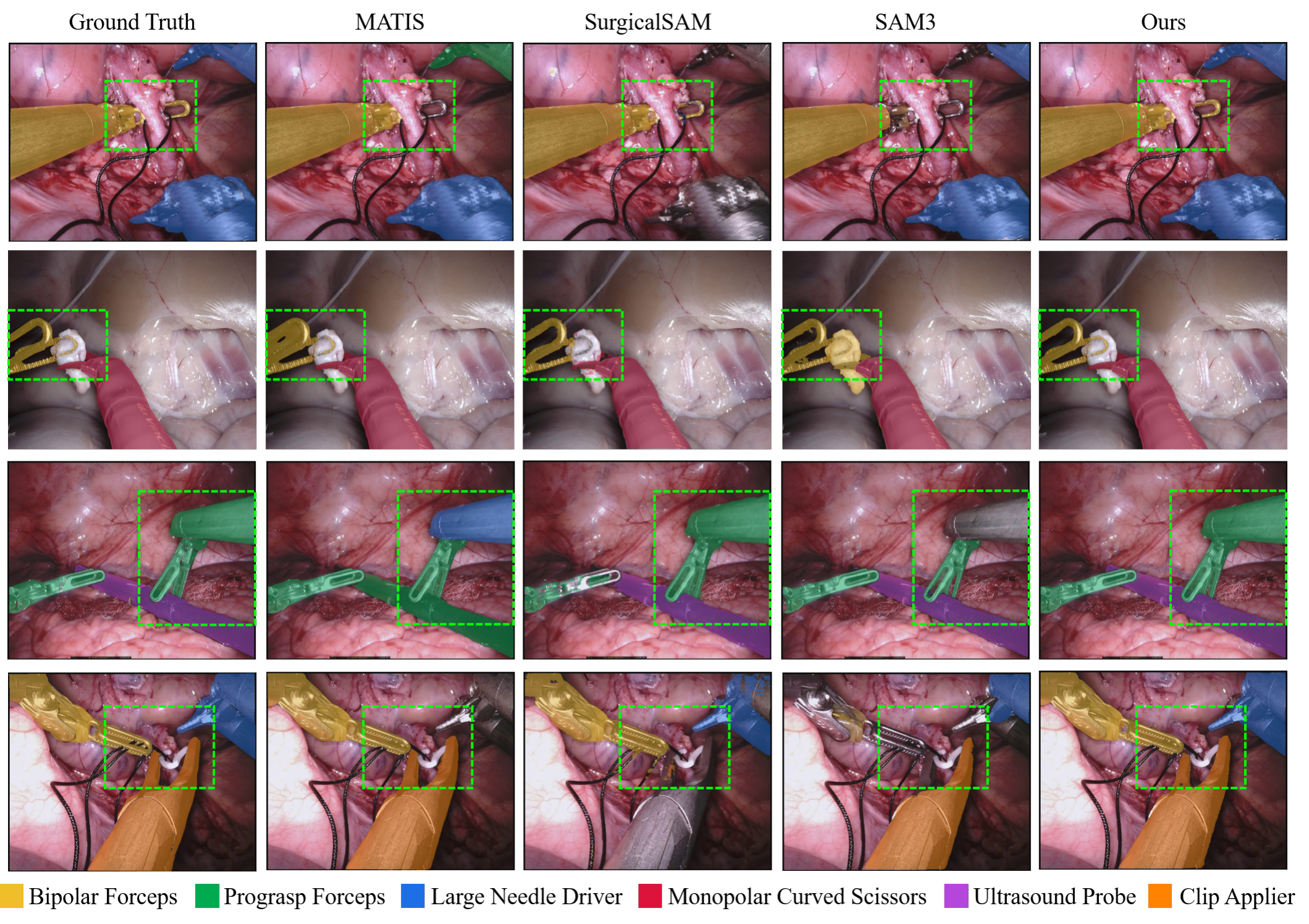}}
\caption{The visual results of the comparative experiments. The green boxes highlight the comparison details.}
\label{fig:visual}
\end{figure*}

\subsection{Implementation Details}

The framework is implemented in PyTorch and trained on NVIDIA A100 GPU. During the adaptation phase, the primary image encoder, language backbone, geometry encoder, and fusion encoder of SAM3 are frozen to retain foundational priors. We selectively unfreeze the final four vision trunk blocks, the final two language encoder blocks, and the final two fusion encoder layers.

The model is optimized with AdamW \cite{loshchilov2018decoupled} using a base learning rate of $8\times{10}^{-5}$, while applying smaller learning rates to the pre-trained vision and language backbones, $2.5\times{10}^{-5}$ and $ 5\times{10}^{-6}$, respectively \cite{bao2022beit}. All input images are resized to $1008\times1008$ \cite{carion2026sam} and trained for 20 epochs. The multi-scale visual prototype memory bank is pre-constructed using K-means clustering on the training set features and K is set to 4. For the training objective, the original SAM3 loss weights are retained, the local alignment loss weight is set to 0.01.

\subsection{Main Results}
We compare HPMA with two groups of representative methods: specialized surgical segmentation models, including ISINet \cite{gonzalez2020isinet}, MATIS Frame \cite{ayobi2023matis}, S3Net \cite{baby2023forks}, SCI-Net \cite{mei2026accurate}, and SAM-based approaches, including TrackAnything \cite{yang2023track}, PerSAM \cite{zhang2024personalize}, SurgicalSAM \cite{yue2024surgicalsam}, MA-SAM2 \cite{yin2025memory}, Distillation-SAM \cite{tang2026distillation}. We additionally report a baseline utilizing ground-truth centroids with SAM3 (GT Centroid + SAM3) \cite{carion2026sam}. Following previous works \cite{yue2024surgicalsam, yue2023surgicalpart, mei2026accurate}, the results for the compared methods are reproduced using their officially released pre-trained weights. However, to ensure a fair comparison and present the strongest baselines, we directly cite the results from the original papers whenever our reproduced performance falls short of their originally reported metrics.

As shown in Table \ref{tab:endovis2017} and Table \ref{tab:endovis2018}, HPMA achieves the best overall performance on both benchmarks. Regarding specialized baselines, although they incorporate carefully designed components, they still face challenges such as temporal dependency or limited data diversity. ISINet aggregates class predictions across frames, but relies on a dedicated detection pipeline and temporal association. S3Net employs multi-scale mask-attended features, but remains dependent on task-specific training with limited surgical data. MATIS introduces masked attention and temporal modeling, yet its category-level performance is constrained, particularly for less frequent instruments. More recently, SCI-Net integrates global-context aggregation, frequency-domain attention, and scale-aware dilation, but it can still be affected by severe reflections and motion blur. In contrast, HPMA explicitly preserves category-specific surgical evidence while retaining the pre-trained knowledge of SAM3, resulting in more balanced performance across instrument categories.

SAM-based methods exhibit strong competitiveness by exploiting foundation-model priors. Notably, the sub-optimal performance of the baseline (GT Centroid + SAM3) reveals that spatial prompts are insufficient without category-level semantics, which explicitly highlights the necessity of injecting stable, category-specific visual evidence. TrackAnything and PerSAM rely on point prompts or reference masks, making their performance sensitive to prompt quality and limiting fully automatic segmentation. MA-SAM2 exploits temporal memory for training-free video segmentation, showing suboptimal performance. SurgicalSAM introduces a prototype-based class prompt encoder and contrastive prototype learning, enabling automatic segmentation without explicit spatial prompts. However, it treats each instrument as a single entity without utilizing the information from different scales. By maintaining a multi-scale prototype memory and coupling hierarchical evidence with appropriate model components, HPMA overcomes these limitations and achieves the best performance among both specialized and SAM-based methods.

Visual comparisons further validate the superiority of HPMA over baseline SAM-based methods. As shown in Fig. \ref{fig:visual}, traditional zero-shot or single-pathway prompt models often fail to precisely delineate boundaries due to the complex visual variations in surgery, such as overlapping tissues. They frequently misclassify the highly similar categories. In contrast, by grounding the adaptation process in our hierarchical visual prototype memory, HPMA maintains robust semantic stability. The scale-matched injection ensures that the model successfully captures holistic semantic meaning while simultaneously refining sharp, localized boundaries, even under severe intraoperative occlusions.

\begin{figure}[!t]
\centerline{\includegraphics[width=1\linewidth]{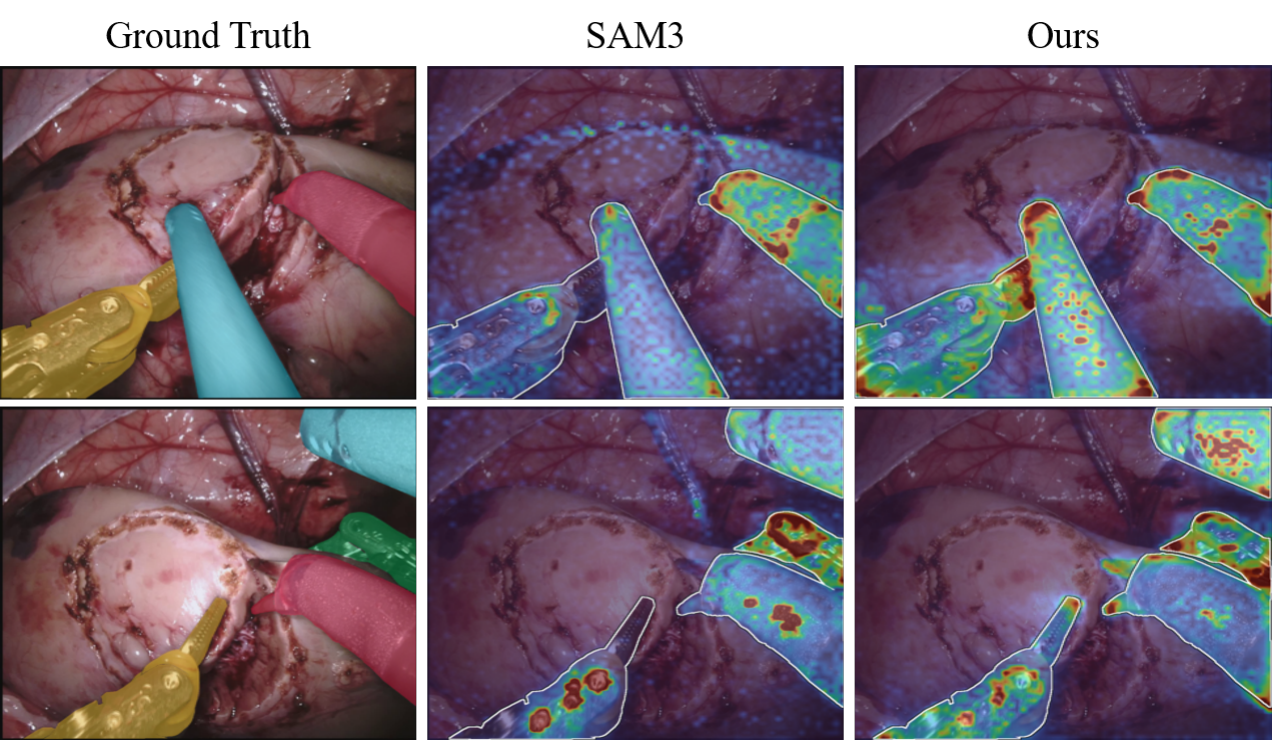}}
\caption{The attention map visualization of SAM3 and our HPMA method.}
\label{fig:heatmap}
\end{figure}

To intuitively demonstrate the effectiveness of our hierarchical coupling mechanism, we visualize the decoder cross-attention maps in Fig. \ref{fig:heatmap}. As shown in the middle column, SAM3 frequently scatters its attention across background tissues. In contrast, after incorporating HPMA (right column), the high-response regions explicitly converge on distinct physical components of the surgical instruments, such as jaws, joints, and shafts. This visual evidence validates that our hierarchical adapters successfully resolve the semantic domain gap and better capture complex instrument structures.

We further evaluate the computational complexity and inference efficiency of HPMA compared to the SurgicalSAM baseline. Specifically, when using the SAM ViT-H backbone, HPMA achieves a lower computational cost of 4917.91 GFLOPs, markedly reducing the computational burden compared to the 5989.78 GFLOPs required by SurgicalSAM. Crucially, this reduction in GFLOPs translates directly to a faster inference speed. HPMA operates at 2.69 FPS, outperforming the 1.98 FPS achieved by SurgicalSAM.

\begin{table}[t]
\centering
\caption{Ablation Study on the EndoVis2018 Dataset.}
\resizebox{0.45\textwidth}{!}{
\begin{tabular}{ccccccc}
\toprule
No. & Global & Structural & Local & Challenge IoU & IoU & mc IoU \\
\midrule
1 & $\times$ & $\times$ & $\times$ & 81.11 & 81.11 & 70.22 \\
2 & $\checkmark$ & $\times$ & $\times$ & 81.90 & 81.90 & 75.38 \\
3 & $\times$ & $\checkmark$ & $\times$ & 82.66 & 82.66 & 75.03 \\
4 & $\times$ & $\times$ & $\checkmark$ & 83.16 & 83.16 & 74.05 \\
5 & $\checkmark$ & $\checkmark$ & $\times$ & 83.44 & 83.44 & 76.89 \\
6 & $\checkmark$ & $\checkmark$ & $\checkmark$ & \textbf{84.12} & \textbf{84.12} & \textbf{77.20} \\
\bottomrule
\end{tabular}
}

\label{tab:ablation}
\end{table}

\begin{table}[t]
\centering
\caption{Ablation Study on the Injection Strategies.}
\resizebox{0.45\textwidth}{!}{
\begin{tabular}{lccc}
\toprule
Injection Strategy & Challenge IoU & IoU & mc IoU \\
\midrule
Multi-scale mean fusion & 82.20 & 82.20 & 74.08 \\
Multi-scale learned fusion & 82.45 & 82.45 & 71.74 \\
G-Text, S-Dec & 83.44 & 83.44 & 76.89 \\
G-Dec, S-Text & 82.60 & 82.60 & 72.50 \\
G-Text, L-SegHead & 82.98 & 82.98 & 76.44 \\
G-Text, S-Dec, L-SegHead & 82.51 & 82.51 & 76.59 \\
G-Text, S-Dec, L-Opt. & \textbf{84.12} & \textbf{84.12} & \textbf{77.20} \\
\bottomrule
\multicolumn{4}{l}{\small \textit{* G refers to Global Level, S refers to Structural Level, L refers to Local Level.}}
\end{tabular}
}

\label{tab:injection_strategies}
\end{table}


\begin{figure}[!t]
\centerline{\includegraphics[width=0.55\linewidth]{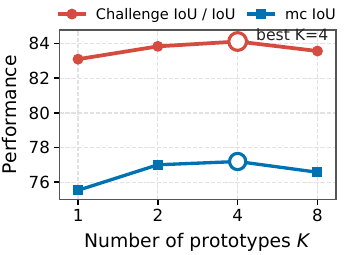}}
\caption{Ablation Study on the value of K.}
\label{fig:ablation_k}
\end{figure}

\subsection{Ablation Study}

To validate our hierarchical coupling mechanism, we conduct an ablation study on the EndoVis2018 dataset. As shown in Table \ref{tab:ablation}, sequentially integrating hierarchical evidence steadily improves performance compared to a partially fine-tuned baseline lacking prototype injection. The addition of global prototypes initially bridges the semantic domain gap. Crucially, incorporating structural-level adapters provides essential geometric priors, substantially boosting the Challenge IoU from 81.90\% to 83.44\%. This confirms that explicit structural bias is vital for parsing highly articulated surgical tools. Finally, integrating the local alignment objective refines high-resolution boundary details, achieving a peak Challenge IoU of 84.12\% and mc IoU of 77.20\%. These progressive gains demonstrate that visual evidence across global, structural, and local levels is complementary and indispensable for robust adaptation.

We further compared different prototype injection strategies. As demonstrated in Table \ref{tab:injection_strategies}, a naive multi-scale fusion yields suboptimal results, indicating that congested prompting pathways create representational bottlenecks. In contrast, routing Global (G) prototypes to the Text features and Structural (S) prototypes to the Decoder significantly outperforms inverted mappings. This validates our core design hypothesis: global features are optimally suited to calibrate abstract semantic classes, while structural features naturally align with the geometric spatial anchors of decoder object queries. Furthermore, directly fusing Local (L) prototypes into the segmentation head degrades performance. Formulating the local level purely as an auxiliary optimization objective circumvents this degradation, yielding the highest Challenge IoU.

Fig. \ref{fig:ablation_k} evaluates the impact of the prototype count $K$ per category. A single prototype ($K=1$) fails to capture the severe intra-class visual variations. Performance peaks at K=4, which is adopted as the optimal configuration.

\section{Conclusion}

We present HPMA, a Hierarchical Prototype-Memory Adaptation framework that robustly adapts the Segment Anything Model 3 (SAM3) for surgical instrument segmentation. By decoupling visual evidence storage into a frozen multi-scale prototype memory bank, HPMA mitigates the prompt degeneration typical of single-pathway methods. Furthermore, our scale-matched adapters seamlessly integrate semantic calibration, structural awareness, and local detail refinement into respective SAM3 modules.  Extensive experiments on the EndoVis2017 and EndoVis2018 benchmarks demonstrate HPMA's state-of-the-art accuracy and robustness against complex intraoperative variations. Building on our computational efficiency over existing baseline, future work will explore model compression and knowledge distillation to transfer these learned capabilities into lightweight backbones. This will facilitate seamless integration with intraoperative platforms, unlocking the potential of foundation models for real-time surgical navigation and robotic assistance. 

{
\bibliographystyle{elsarticle-num}

\bibliography{aaai2027}
}


\end{document}